\documentclass[11pt,a4paper]{article}
\usepackage[hyperref]{acl2021}
\usepackage{times}
\usepackage{graphicx}
\usepackage{latexsym}

\usepackage{microtype}
\usepackage{hyperref}

\aclfinalcopy 
\def\aclpaperid{87} 

\title{What's in the Box? \\ A Preliminary Analysis of\\ Undesirable Content in the Common Crawl Corpus}

\author{Alexandra (Sasha) Luccioni \\
  Université de Montréal \& \\
  Mila Québec AI Institute\\
  \texttt{sasha.luccioni@mila.quebec} \\\And
 Joseph D. Viviano\\
 Mila Québec AI Institute\\
  \texttt{joseph@viviano.ca} \\}

\date{}

\begin{document}
\maketitle

\begin{abstract}
Whereas much of the success of the current generation of neural language models has been driven by increasingly large training corpora, relatively little research has been dedicated to analyzing these massive sources of textual data. In this exploratory analysis, we delve deeper into the Common Crawl, a colossal web corpus that is extensively used for training language models. We find that it contains a significant amount of undesirable content, including hate speech and sexually explicit content, even after filtering procedures. We discuss the potential impacts of this content on language models and conclude with future research directions and a more mindful approach to corpus collection and analysis.
\end{abstract}

\section{Introduction}
In recent years, much of the progress in Natural Language Processing (NLP) research has been largely driven by Transformer-based language models, which have pushed forward the state-of-the-art in tasks such as question answering~\cite{rajpurkar2018know} and natural language inference~\cite{bowman2015large}. However, these increasingly complex models also require increasingly large amounts of data to train them, which is often a combination of curated, high-quality datasets such as encyclopedic articles and books and non-curated content from the Web~\cite{radford2018improving, radford2019language}. This second category of large, non-curated dataset is becoming increasingly popular as they are required to train large language models. 

The current largest dataset used for training neural language models, the \href{https://commoncrawl.org/}{Common Crawl}, is a non-curated corpus consisting of multilingual snapshots of the web. New versions of the Common Crawl are released monthly, with each version containing 200 to 300 TB of textual content scraped via automatic web crawling. This dwarfs other commonly used corpora such as \href{https://en.wikipedia.org/}{English-language Wikipedia}, which adds up to roughly 5.6 TB of data, and the BookCorpus, which only represents around 6 GB~\cite{zhu2015aligning}. The Common Crawl has been used to train many of the recent neural language models in recent years, including the GPT model series~\cite{radford2018improving,brown2020language}, BERT~\cite{devlin2018bert} and FastText~\cite{grave2018learning} and, given its size, often represents the majority of data used to train these architectures. 

In the current article, we present an initial analysis of the Common Crawl, highlighting the presence of several types of explicit and abusive content even after filtering. We discuss our findings and, given the potential downstream impact of this content on language models, we discuss the importance of ensuring that the corpora we use for training language models are extracted more mindfully and with more emphasis on their quality and propose avenues of research to achieve this goal. 

\section{Related Work}

In recent years, a growing body of research in NLP has unearthed biases in common language models~\citep{bolukbasi2016man,sheng2019woman, zhao2019gender,bordia2019identifying, hutchinson2020social}.
This work has raised important questions regarding the impact of these embedded biases on downstream decision-making, given the increasing usage of these models in various applications. Consequently, much work has also been dedicated to creating standardized diagnostic tests to detect these biases~\cite{caliskan2017weat,may2019measuring, nadeem2020stereoset, sweeney2019transparent} and to remove them~\cite{bolukbasi2016man,zhao2018learning, manzini2019black}, although the extent to which this is possible is still under debate~\cite{gonen2019lipstick}. In fact, research has found that \emph{``The biases found in Internet-scale language models like GPT-2 are representative of the data on which the model was trained''}~\cite{solaiman2019release}, which can be directly linked to the presence of hate speech on the Internet~\cite{abid2021persistent}.

However, given the importance of this research, comparatively little attention has been dedicated to analyzing the corpora used to train language models. This is understandable because frequently used datasets such as the Common Crawl contain truly massive amounts of data, making it challenging to mine it for meaningful insights. In fact, a recent survey on automatic web page classification has deemed the task difficult not only due to the complexity and heterogeneity of web content, but also due its the high computational cost, suggesting that machine learning (ML) approaches have much to contribute to it~\cite{hashemi2020web}. While certain notable endeavors have indeed analyzed specific aspects of corpora such as the Common Crawl~\cite{kolias2014exploratory, caswell2021quality} and Wikipedia~\cite{hube2017bias}, they have only scratched the surface of what these bodies of text contain. For instance, recent work has found that the Common Crawl contained over 300,000 documents from unreliable news sites and banned subReddit pages containing hate speech and racism~\cite{gehman2020realtoxicityprompts}, while complementary research has shown that individual training examples can be extracted by querying language models~\cite{carlini2020extracting}, together illustrating that the presence of questionable content is a significant issue for statistical language models. In the current work, we endeavor to understand the content and quality of the Common Crawl as a first step towards establishing more consistent approaches to filtering and refining it.

\section{Analyzing the Common Crawl}

Given its size, both downloading and analyzing the Common Crawl are time-consuming and costly endeavors. The \href{https://commoncrawl.org/2020/12/nov-dec-2020-crawl-archive-now-available/}{most recent version} of the Common Crawl, dating from November/December 2020, has 2.6 billion web pages in raw text format, saved in `shards' each containing of tens of thousands of pages. Given our hardware constraints, we chose to focus on a subset of the corpus, randomly sampling 1\% of the files it contains, which after filtering by language amounts to roughly 115 GB of textual content or 5,835,339 web pages in total, which we analyzed in terms of hate speech, adult content, and efficacy of perplexity-based filtering~\footnote{All code used in these analysis are publicly available: https://github.com/josephdviviano/whatsinthebox}. In this work, we focus on detecting sexually-explicit and hate speech, since they represent common examples of ``undesirable" content that can be generally seen as inappropriate for a language model to generate in most situations. We acknowledge that desirable model behaviour is application specific, and believe our findings can extend to any other ``undesirable" topic that might be present in available language corpora.  We present our results in the sections below.

\subsection{Detecting Hate Speech}
The existence of hate speech on the internet has been described as ``an important societal problem of our time", with ``profound and lasting" psychological effects on its victims~\cite{mishra2019tackling}. As such, a substantial amount of NLP research dedicated to automating hate speech detection, with several datasets and approaches being proposed in recent years~\cite{schmidt2017survey, mishra2019tackling, vidgen2020directions, kiritchenko2018examining}. Most of this research is carried out on data extracted from social media sources such as Twitter~\cite{founta2018large, basile2019semeval, waseem2016hateful} and Reddit~\cite{tadesse2019detection, farrell2019exploring}, with both ML-based~\cite{badjatiya2017deep} and count-based approaches~\cite{davidson2017automated} achieving comparable results~\cite{fortuna2018survey}. In order to estimate the quantity of hate speech in the Common Crawl, we endeavored to compare 3 approaches: DELIMIT, a recent BERT-based model trained on social media data~\cite{aluru2020deep}, Hate Sonar, a Logistic Regression approach trained on data from Web fora and Twitter~\cite{davidson2017automated} and a n-gram-based approach using a list of n-grams extracted from \href{https://hatebase.org/}{Hate Base}. We present samples of text flagged by all of these approaches in Table~\ref{hateresults}, below. 

\begin{table}[h!]

\begin{tabular}{p{1.3cm}|p{5cm}}
\multicolumn{1}{c|}{\textbf{Approach}} & \multicolumn{1}{c}{\textbf{Text}} \\ \hline
\textbf{HateSonar} &
  \begin{tabular}[c]{@{}l@{}}  \small Their US/Euro plan put in your face:\vspace{-0.1cm} \\  \small  demonic jews hate white goyim!\\   \small Such sick and twisted people, white \vspace{-0.1cm}\\  \small people are.\end{tabular} \\ \hline
\textbf{Delimit}                       &     \begin{tabular}[c]{@{}l@{}} 
\small they are only stupid arab from wp-ar haha \\ 
\small Yeah, dumb ass n*gger $\dagger$ \\
\end{tabular} \\ \hline
\textbf{N-gram}                        &
\small nude attention whore asian bastards\\ &
\small  In America all male look like this homo \\

\end{tabular}
\caption{Examples of hate speech found by the approaches tested. Examples with $\dagger$ have been censored by the authors.}
\label{hateresults}
\end{table}

We found that the three approaches compared suggest similar proportions of websites containing hate speech : 5.24\% of websites from our sample were flagged by DELIMIT, 4.02\% by HateSonar, and 6.38\% by the n-gram approach~\footnote{We are conscious of the high false positive rate of n-gram approaches and therefore only consider sites to be flagged if they contain 3 or more n-grams from the list.}. Qualitative analysis of a sample of sites flagged by each approach showed that while n-grams picked up on racial slurs, HateSonar also detected debates about racial supremacy and racially-charged conspiracy theories. Many of the sites that DELIMIT flagged were adult content with mentions of violent acts towards specific ethnic groups, illustrating the fine line between sexual violence and hate speech, which we elaborate further in the following subsection. Generally speaking, the presence of even a small fraction of websites that incite hate in training corpora is worrisome since it can result in models that replicate this kind of discourse when prompted~\cite{wolf2017we, carlini2020extracting}. 

\subsection{Sexually Explicit Content}

Compared to hate speech, the detection of sexually explicit content has received less attention from the NLP community, with existing ML approaches focusing mainly on the detection of explicit images~\cite{wehrmann2018adult,rowley2006large} and URLs~\cite{matic2020identifying}, whereas n-gram-based approaches remain predominantly used in practice by web providers~\cite{hammami2003webguard, polpinij2006content, ho2004statistical}. In our analysis, we used a \href{https://github.com/IDEA-NTHU-Taiwan/porn_ngram_filter}{list of n-grams} extracted from adult websites in order to establish the percentage of websites from our sample that contained sexually explicit content; however, we found no available statistical or ML-based approach that we could use to compare our count-based approach with. The n-gram approach detected that 2.36\% of the web pages that we analyzed contained at least one of the words from our list, with 1.36\% containing 3 or more and 0.73\% containing 10 or more (see Table~\ref{perplexity} for results). We show a sample of the URLs flagged by our approach in Table~\ref{sexual}, below.

\begin{table}[h!]
\begin{tabular}{p{7cm}}
\multicolumn{1}{c}{\textbf{Page URL (\texttt{http://} removed)}}\\ %
 \hline
 \small \texttt{adultmovietop100.com/}\\
 \small \texttt{erohon.me/}\\
 \small \texttt{celebrityfan.net/}\\
 \small \texttt{queantube.com/}\\
 \small \texttt{adelaide-femaleescorts.webcam}\\
  \hline
\end{tabular}
\caption{Sample of URLs of adult content websites identified by the n-gram approach. Protocol removed to prevent URL generation.}
\label{sexual}
\end{table}

While a few percent of sexually explicit content may not seem like much, the type of language and content contained on adult websites can have harmful repercussions. For instance, the prevalence of sexual violence towards women, especially towards women of color, on adult websites~\cite{foubert2019pornography, shim2015analysis, fritz2020worse} may contribute to further dissemination and amplification of these biases in downstream models. As modern language models have no way to evaluate generation appropriateness, models trained with even a small proportion of these undesirable inputs cannot be guaranteed to avoid generating outputs with similar biases if presented with a specific context or prompt. This is a risk that is important to mitigate in applications, where the general-purpose language models can end up being used in applications used by sensitive groups in professional contexts or minors, such as chatbots and toys. 


\subsection{Filtering by Perplexity Score}
While the analyses described above were carried out on unfiltered web pages from the Common Crawl, the training pipeline of many large-scale NLP models involves some type of filtering and cleaning, from excluding low-quality content~\cite{grave2018learning} to fuzzy deduplication~\cite{brown2020language}. 
One such popular filtering approach is based on training a language model on a target, high-quality domain such as Wikipedia, and using it to calculate the perplexity score of web pages using this model~\cite{wenzek2020ccnet}. To test the efficacy of this scoring procedure, we calculated the perplexity score of each web page from our sample of the Common Crawl and used it to separate pages into 3 equal buckets (high, middle and low-quality) based on their perplexity. We compare the percentages of hate speech and sexually explicit content for the entire sample, as well as the high- and low-quality documents, in Table~\ref{perplexity}. 

\begin{table}[ht]
\centering

\begin{tabular}{l|l|l|l}
 &
  \textbf{\begin{tabular}[c]{@{}l@{}}Entire\\ Sample\end{tabular}} &
  \textbf{\begin{tabular}[c]{@{}l@{}}High \\ Quality\end{tabular}} &
  \textbf{\begin{tabular}[c]{@{}l@{}}Low \\ Quality \end{tabular}} \\ \hline
\textbf{\begin{tabular}[c]{@{}l@{}}1+ sexual \\ n-grams\end{tabular}}      & 2.36\%  & 1.81\%  & 3.97\%  \\ \hline
\textbf{\begin{tabular}[c]{@{}l@{}}3+ sexual \\ n-grams\end{tabular}}      & 1.36\%  & 0.42\%   & 3.11\%  \\ \hline
\textbf{\begin{tabular}[c]{@{}l@{}}10+ sexual \\ n-grams\end{tabular}}     & 0.73\%  & 0.08\%   & 1.98\%  \\ \hline
\textbf{\begin{tabular}[c]{@{}l@{}}1+ hate \\ n-grams\end{tabular}}        & 17.78\% & 18.95\% & 17.19\% \\ \hline
\textbf{\begin{tabular}[c]{@{}l@{}}3+hate \\ n-grams\end{tabular}}         & 6.38\%  & 6.19\%   & 8.26\%  \\ \hline
\textbf{\begin{tabular}[c]{@{}l@{}}10+ hate \\ n-grams\end{tabular}}       & 1.16\%  & 1.17\%   & 1.70\%  \\ \hline
\textbf{\begin{tabular}[c]{@{}l@{}}Hate speech \\ (Sonar)\end{tabular}}   & 4.02\%  & 3.47\%   & 5.09\%  \\ \hline
\textbf{\begin{tabular}[c]{@{}l@{}}Hate speech \\ (Delimit)\end{tabular}} & 5.24\%  & 5.77\%   & 5.66\% 
\end{tabular}
\caption{Comparison of hate speech and sexual content detected in the entire corpus, as well as high- and low- quality sites.}
\label{perplexity}
\end{table}

\begin{figure}
    \centering
    \hspace*{-20pt}\includegraphics[width=0.45\textwidth]{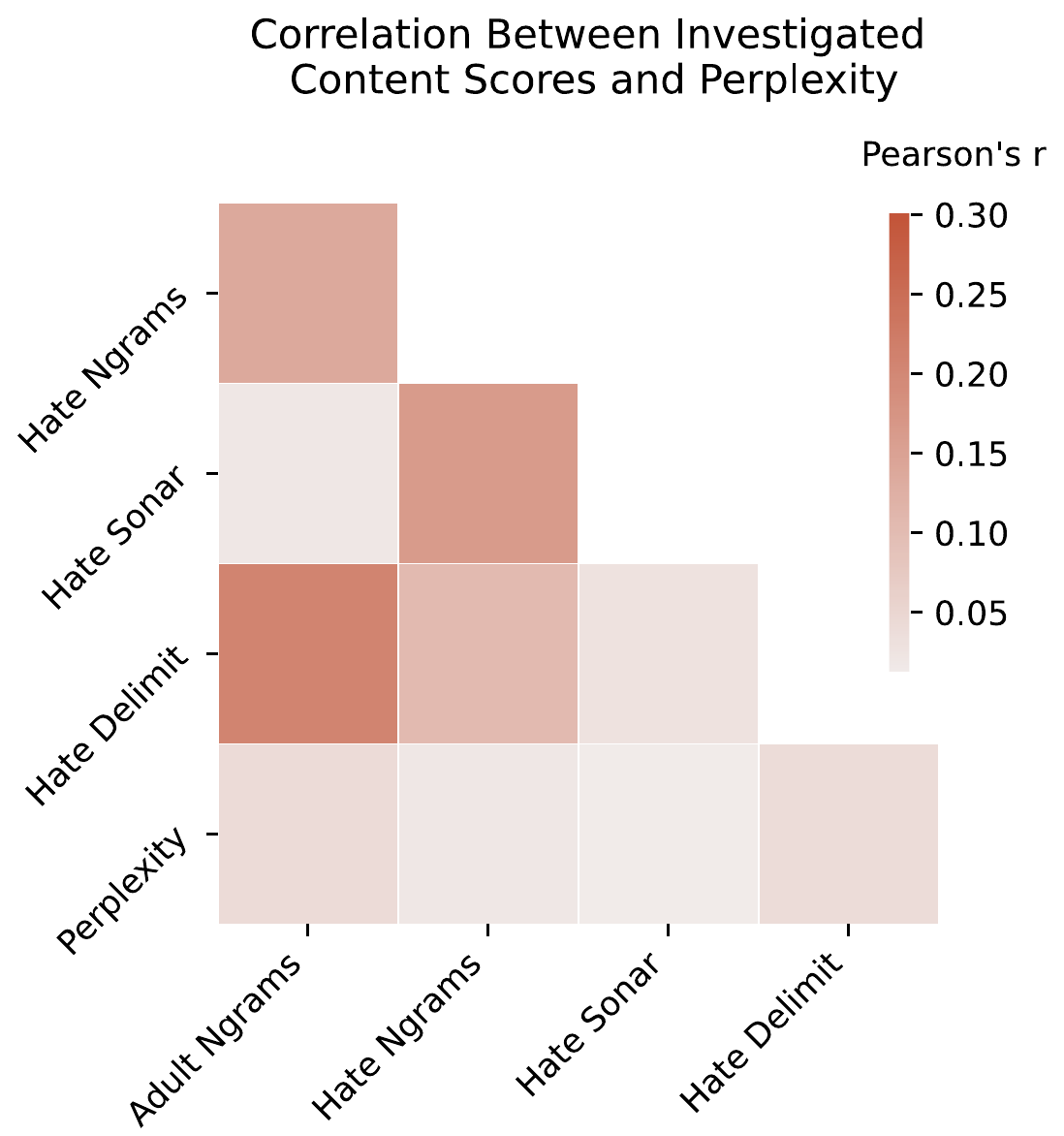}
    \caption{Correlation coefficients (Pearson's \textit{r}) calculated between all content metrics investigated and perplexity, a commonly-used text quality metric.}
    \label{fig:correlations}
    \vspace{-10pt}
\end{figure}

While filtering by perplexity does seem to filter out many websites containing sexual content, it does not detect much of the hate speech that is flagged by the count-based or statistical methods. In fact, perplexity scores had low correlations with all detection methods tested (Figure \ref{fig:correlations}). This supports the methodology of Wenzek et al.~\citeyearpar{wenzek2020ccnet}, who noted that while
\emph{``perplexity was a relative good proxy for quality"}, also argued that some of the lower-quality texts could still be useful for specific applications, and therefore did not use it to exclude documents from the training set of their language model. While we are exploring ways of modifying the original approach in order to be more discerning, we believe that there more nuanced metrics that can be used for estimating and filtering documents based on text, potentially coupling embedding-based approaches with statistical ones.

\subsection{Behaviour of Different Detection Methods}

The approaches that we compared in the current study are different in the features that they use and techniques employed for detecting particular types of content. HateSonar employs classical NLP techniques for hate speech detection, constructing features from Penn Part-of-Speech N-grams with TF-IDF weighting based on a hand-crafted hate speech dataset, training simple classifier ensembles using Support Vector Machines, random forests, naive Bayes, and linear models. Delimit, on the other hand, is A BERT-based model trained on Twitter and Reddit posts, not relying on any handcrafted features. Our simple n-gram approach unsurprisingly was more in agreement with HateSonar than Delimit, given that both rely on count-based features. The fact that all methods identified different instances of clear hate speech implies that we are far from a general purpose dataset-filtering approach. These results also imply that deep learning models learn very different features to classify hate speech than other methods, and given their sensitivity to the specific composition of the dataset used to train them (as exposed by the propensity of large models to memorize training examples \cite{carlini2020extracting}), the presence of undesirable content in the corpora used to train them should be taken seriously.

\section{Discussion}

\subsection{Summary of Results}
We recognize that the exploratory work presented above is only the tip of the iceberg in terms of the analyses that can be done on the massive web corpora that are feeding our language models. However, analyzing the Common Crawl would require computational resources far in excess of what is available to most research institutions.  We therefore hope that this initial analysis will inspire our fellow researchers to continue to dig deeper into this topic, and to propose more scalable, thorough, and nuanced approaches for analyzing the massive corpora used to train language models. We also recognize this analysis would have been more comprehensive on a small curated dataset, but given the amount of data needed to train modern language models, we believe the community needs to move beyond analysis techniques only compatible with small-data, toward something that will scale to the datasets used to train these large models.

Also, while we have currently adopted a purely descriptive approach, we feel that it is worth discussing and debating the consequences of our analysis, and those of our peers, within the NLP community. While it can be argued that the Common Crawl corpus is an accurate portrayal of the discourse of modern society -- which includes sexual content, hate speech, and racial and gender biases -- we believe that it is up for debate whether this discourse is the one that we, as a community, want to use to train the models that translate our texts, influence our search results and answer our questions. Notably, the Common Crawl over-represents those populations that are avid users of the internet: younger, English-speaking individuals from developed countries, who are those who have the most access to the internet globally~\cite{WorldBankInternet}. Furthermore, internet communities supported by anonymity and and particular norms can amplify toxic discourse that would not be found in mainstream corpora \cite{massanari2017gamergate} often exacerbated by the well-documented 'online disinhibition' phenomenon where users find themselves more likely to engage in anti-social behaviours due to the lack of immediate social feedback \cite{wachs2019understanding, mathew2019temporal, de2021characterizing}. This can further perpetuate the lack of diverse, representative language models that can adequately mirror society beyond the boundaries of internet communities.

\subsection{Future Work}

Given the general superior performance of large language models on common benchmarks, and that they require ever larger datasets to train them, we believe it is important that for the ML community to carry out a more extensive analysis of: 1) the impact of undesirable content in the datasets used to train these models on downstream performance;  2) the effect of properly filtering these examples out of the dataset \textit{before} model training, and 3) approaches for regularizing model outputs to be acceptable regardless of the data used to train the model. All three directions require a better understanding of the contents of the datasets, which we believe requires new tools that are scalable to the Common Crawl (or similarly large and diverse corpora) to identify such examples. Models trained to detect undesirable examples, like the ones used in this paper, need to be improved such that they can reliably generalize to the Common Crawl, which constitutes a significant undertaking. Additionally, future work could explore the utility of controlling model generation using labelled ``undesirable" examples \cite{zhang2020pointer,engel2017latent}, or human-in-the-loop learning methods \cite{wang2021putting} for fine-tuning a language model trained using undesirable examples. It will also be important to evaluate whether curation is sufficient: it remains possible that a model could create an undesirable generation from multiple distinct innocuous examples~\cite{bender2021dangers,gehman2020realtoxicityprompts}. It is also worth considering that for some applications, task-focused models with curated training examples may perform better than large models trained on unfiltered corpora, so that their behaviour can be more reliably guaranteed: these are all interesting avenues for future work.

Finally, while larger corpora generally result in better models~\cite{kaplan2020scaling,sun2017revisiting}, data quality and corpora content also plays a major role in the caliber and appropriateness of these models for the various downstream applications~\cite{florez2019unintended,abid2021persistent,bhardwaj2021investigating}. To produce high quality and safe neural language models will likely require the community to adopt more mindful data collection practices ~\cite{gehman2020realtoxicityprompts,bender2018data,gebru2018datasheets,jo2020lessons,paullada2020data, bender2021dangers}, establish standardized filtering pipelines for corpora~\cite{roziewski2016languagecrawl,OrtizSuarezSagotRomary2019, wenzek2020ccnet}, and develop methods for evaluating the bias in trained models \cite{schick2021self}. We recognize that this is not a straightforward task with a one-size-fits all solution, but we propose that as much attention should be dedicated to the corpora used for training language models as to the models themselves, and that corpora transparency is a prerequisite for language model accountability. 

\clearpage
\bibliography{acl2020}

\end{document}